\pdfoutput=1
\documentclass[a4paper,11pt]{article}
\usepackage[utf8]{inputenc}
\usepackage[T1]{fontenc}    
\usepackage[font=small,skip=1pt]{caption}
\usepackage[rmargin = 2.5cm, lmargin = 2.5cm, tmargin = 2.5cm, bmargin = 2.5cm]{geometry}
\usepackage{authblk}
\PassOptionsToPackage{numbers,sort&compress}{natbib}
\usepackage{natbib}

\usepackage{enumerate} 
\usepackage{enumitem} 

\usepackage{graphicx}

\usepackage{amsmath}
\usepackage{amssymb}
\usepackage{amsthm}

\usepackage{xr-hyper}
\usepackage{hyperref}       
\usepackage{url}            
\usepackage{booktabs}       
\usepackage{amsfonts}       
\usepackage{nicefrac}       
\usepackage{microtype}      
\usepackage{xcolor}         
\usepackage{setspace} 

\def    \be            {\begin{equation}}
\def    \ee            {\end{equation}}
\def    \bea           {\begin{eqnarray}}
\def    \eea           {\end{eqnarray}}

\theoremstyle{plain}
\newtheorem{theorem}{Theorem}[section]

\theoremstyle{definition}

\theoremstyle{remark}

\def    \be            {\begin{equation}}
\def    \ee            {\end{equation}}
\def    \bea           {\begin{eqnarray}}
\def    \eea           {\end{eqnarray}}
\def    \ba            {\begin{align}}
\def    \ea            {\end{align}}

\DeclareMathOperator*{\argmin}{arg\,min}

\title{
A general Markov decision process formalism for action-state entropy-regularized reward maximization
} 

\author[1]{Dmytro Grytskyy}
\author[1]{Jorge Ramírez-Ruiz}
\author[1,2]{Rubén Moreno-Bote}

\affil[1]{Center for Brain and Cognition, and Department of Information and Communication Technologies, Universitat Pompeu Fabra, Barcelona, Spain}
\affil[2]{Serra Húnter Fellow Programme, Universitat Pompeu Fabra, Barcelona, Spain}
\date{\today}

\begin{document}

\maketitle

\begin{abstract}
Previous work has separately addressed different forms of action, state and action-state entropy regularization, pure exploration and space occupation.
These problems have become extremely relevant for regularization, generalization, speeding up learning and providing robust solutions at unprecedented levels.
However, solutions of those problems are hectic, ranging from convex and non-convex optimization, and unconstrained optimization to constrained optimization. 
Here we provide a general dual function formalism that transforms the constrained optimization problem into an unconstrained convex one for any mixture of action and state entropies. 
The cases with pure action entropy and pure state entropy are understood as limits of the mixture. 
\end{abstract}

\section{Introduction}

It is well known that classical reinforcement learning, understood as learning from external rewards, has severe limitations. While it has been posited that reward is ``enough" to learn any behavior \cite{silver2021reward}, agents interacting with the real world often have only access to sparse rewards. Many approaches have been proposed to overcome the sparse reward limitation, endowing agents with additional signals to be optimized along with the rewards. These include minimizing surprise by refining predictions \cite{achiam2017surprise,pathak2017curiosity,burda2018exploration,pathak18largescale,fountas2020deep,hafner2020action}, novelty seeking by visiting states with low visit counts \cite{bellemare2016unifying,tang2017exploration,aubret2022information}, generating actions that leads to predictable transitions (empowerment) \cite{klyubin2005empowerment,jung2011empowerment,mohamed2015variational}, or seeking pure state entropy \cite{hazan2019provably} and related forms of pure exploration objectives \cite{pathak2017curiosity,lee2019efficient,jin2020reward,zhang2021exploration,mutti2021task,eysenbach2021maximum}, to name a few. 

A popular choice for augmenting the reward signal --the one that we focus on in this paper-- is with entropy regularization \cite{todorov2009efficient,ziebart2010modeling,tishby2011information,schulman2017equivalence,nachum2017bridging,haarnoja2018soft,hausman2018learning,eysenbach2018diversity,galashov2019information}. The idea is that the agent will be driven, all else equal, to visit states and taking actions that make the agent act as random as possible (pure entropy regularization, e.g., \cite{haarnoja2018soft}) or penalize the agent for having a policy very different from a default policy (KL regularization, e.g., \cite{todorov2009efficient}). Using this type of regularization can lead to better exploration \cite{hazan2019provably}, more variable and realistic behaviors \cite{ramirez2022seeking}, more efficient learning \cite{schulman2015trust,haarnoja2018soft} and more robust solutions \cite{ziebart2010modeling} against noise and adversarial attacks \cite{eysenbach2021maximum} than classical reinforcement learning algorithms.

While the above approaches use entropy as a regularizer to the optimization reward problem, the specific type of entropy regularizer varies widely across studies, and as a result the approaches and the solutions are hectic. For instance, some use pure action entropy regularization \cite{neu2017unified,nachum2017bridging,haarnoja2018soft,hausman2018learning}, others employ purely state entropy \cite{hazan2019provably}, others take advantage of KL action regularization \cite{todorov2006linearly,schulman2017equivalence,galashov2019information}, and yet others combine action and state pure entropy in balanced \cite{peters2010relative,tishby2011information} or arbitrary ways \cite{ramirez2022seeking}. The variety of approaches and techniques makes it hard to know what the specific effects of different weighting of action and state entropies are on exploration, robustness, and generalization.

To address these questions, it is important to first develop an overarching theory that ideally includes all previous approaches as limits or special cases. This is indeed the goal and achievement of this paper. We first show that augmenting the standard reward objective with a non-negative mixture of action and state entropies leads to a convex optimization problem. We next show that the optimization problem can be cast in its dual form, from where it is easier to prove existence and uniqueness of the optimal distribution $p(s,a)$ over states $s$ and actions $a$. 
Here we focus on the harder problem where the future reward discount factor equals one, $\gamma=1$. The case $\gamma < 1$ can be treated using more standard approaches. As expected, KL regularization is a special case where rewards are simply redefined. 
Next, we derive a general iterative method to find the optimal solution, which is shown to converge in practice. 
We conclude by showing in simple examples how the optimal behavior of the agent depends on the values of the action-state entropy mixture, which suggests that it can be a flexible way for biasing exploration towards actions or states.

\section{Results}

\subsection{Markov decision problem with a mixture of action entropy and state entropy}

The agent is modeled as a finite action-state Markov decision process (MDP) in discrete time. At any state $s$ the agent can choose one action $a$ out of several possible (non-empty set) with a probability $\pi(a|s)$. 
Given the state $s$ and action performed $a$, a new state $s'$ results with probability $p(s'|s,a)$. 
The set of all probabilities $\pi = \{\pi(a|s)\}$ defines the agent's policy. 
We do not need to assume that the MDP be ergodic, and thus different subsets of states can be disconnected and absorbing states are allowed. 
The distribution over states and actions $p(s,a)$ is the stationary probability under the policy, understood as frequency of visiting $(s,a)$ averaged over time, realizations over transitions and initial conditions following the same distribution. 

We define the immediate policy-dependent reward for being at state $s$ and performing action $a$ as 
\be
R_{\pi}(s,a) =
r(s,a) -\alpha \log \pi(a|s) - \beta \log p_{\pi}(s)
\;.
\label{eq_reward}
\ee
\noindent
The first term on the right-hand side is the standard policy-independent reward $r(s,a)$ (assumed to be finite), while the second and third terms correspond to policy-dependent intrinsic rewards \cite{hazan2019provably,schulman2017equivalence,haarnoja2018soft,ramirez2022seeking}. The first one is an exploration bonus for performing rare actions, while the second is an exploration bonus for visiting rare states. The parameters $\alpha \geq 0$ and $\beta \geq 0$ measure the relative strengths between action and state entropies and with respect to the reward. 
Note that it is possible to introduce additional terms of the form $\log \pi_0(a|s)$ and $\log p_0(s)$, where $\pi_0(a|s)$ is an arbitrary default policy and $p_0(s)$ is an arbitrary default stationary distribution by replacing the log terms in Eq. (\ref{eq_reward}) by $-\alpha \log(\pi(a|s)/\pi_0(a|s))-\beta \log(p(a|s)/p_0(a|s))$; introducing a default policy ($\alpha=1$ and $\beta=0$) is customary in approaches where the entropy regularization is replaced by a KL regularization \cite{todorov2009efficient,schulman2017equivalence,galashov2019information}. The formalism and solutions that we describe below do not change, and the above case can be simply obtained by replacing in every equation $r(s,a)$ with the policy-independent total reward $\tilde{r}(s,a) = r(s,a) + \alpha \log \pi_0(a|s) + \beta \log p_0(s)$. The additional terms can be understood as generating an additional form of intrinsic reward, which can be arbitrarily negative for impossible actions or unreachable states under the default policy and default state probability distribution. 

We define the average total reward of Eq. (\ref{eq_reward}) as
\be
R_{\pi} = \sum_{s,a} p_{\pi}(s,a) 
\left(r(s,a) - \log( \pi^{\alpha}(a|s) p_{\pi}^{\beta}(s)   )
\right)
\;,
\label{eq_average_reward}
\ee
\noindent
where $p_{\pi}(s,a) \equiv \pi(a|s) p_{\pi}(s)$ is the stationary joint state-action distribution under policy $\pi(a|s)$ (existence a stationary distribution is not strictly required, that is, solutions can be periodic, and then $p_{\pi}(s,a)$ is understood as action-state frequency counts over the long run) 
The objective is to find the policy $\pi(a|s)$ that maximizes average total reward
\be
R_{max} = \max_{\pi} \sum_{s,a} p_{\pi}(s,a) 
\left(r(s,a) - \log( \pi^{\alpha}(a|s) p^{\beta}_{\pi}(s)   )
\right)
\;,
\label{eq_average_reward_max_pi}
\ee
\noindent
under suitable constraints (see below). However, this objective is in general non-convex in $\pi$ (e.g., for $\alpha=0$ and $\beta=1$, see \cite{hazan2019provably}). 

Eq. (\ref{eq_average_reward}) includes, and expands, a number of standard problems that have been considered previously: action-state entropy $(\alpha,\beta)=(1,1)$ \cite{peters2010relative}, pure state-dependent action entropy $(\alpha,\beta)=(1,0)$ \cite{neu2017unified}, and pure state entropy $(\alpha,\beta)=(0,1)$ \cite{hazan2019provably}. Action-state entropy problems \cite{peters2010relative} leads to an unconstrained optimization problem through the dual function approach; pure state-dependent action entropy problems leads to a constrained convex optimization problem \cite{neu2017unified}; pure state entropy problems leads to a convex optimization problem \cite{hazan2019provably}.

The main contribution of this paper is to show that the average total reward optimization ({\em primal}) problem defined in Eq. (\ref{eq_average_reward}) can be transformed into a convex optimization problem for $\alpha \geq 0$ and $\beta \geq 0$ that can be easily solved as unconstrained convex optimization using the {\em dual} problem for $\alpha>0$ and $\beta>0$. The pure cases with $\alpha=0$ or $\beta=0$ are understood as limits of the obtained solutions.

\subsection{The optimization problem is concave in $p(s,a)$}

The optimization problem can be recast as a convex optimization problem by optimizing $p(s,a)$ instead of the policy $\pi(a|s)$. Once the optimal $p(s,a)$ is obtained, the state-distribution is defined as $p(s)=\sum_a p(s,a)$, and the policy is realized as $\pi(a|s) = p(s,a)/p(s)$, where we assume that $p(s)>0$. For $s$ such that $p(s)=0$, the policy $\pi(a|s)$ can be arbitrarily defined. Written as a function of $p = \{p(s,a)\}$, the average total reward is
\be
R_{p} = \sum_{s,a} p(s,a) 
\left(r(s,a) - \log \left( \frac{p^{\alpha}(s,a)}{p^{\alpha}(s)} p^{\beta}(s)   \right)
\right)
\;.
\label{eq_average_reward_p}
\ee
\noindent
The objective is to maximize the average total reward with respect to $p$,
\be
R_{max} = \max_p \sum_{s,a} p(s,a) 
\left(r(s,a) - \log \left( \frac{p^{\alpha}(s,a)}{p^{\alpha}(s)} p^{\beta}(s)   \right)
\right)
\label{eq_average_reward_max_p}
\ee
\noindent
under the constraints
\bea
&& p(s,a) \geq 0 
\label{eq_cond1}
\\
&& \sum_{s,a} p(s,a) = 1
\label{eq_cond2}
\\
&& \sum_{s,a} p(s'|s,a) p(s,a) = \sum_b p(s',b)
\label{eq_cond3}
\;.
\eea
\noindent
The expression $\sum_a g(s,a)$ denotes the sum over all actions available at state $s$. 

\begin{theorem}[]
	The average total reward $R_p$ is concave in $p$ for $\alpha \geq 0$ and $\beta \geq 0$.
	\label{th_concave_1}
\end{theorem}

\noindent
To see this, we first rewrite $R_p$ as
\bea
R_p 
&=&
\sum_{s,a} p(s,a) r(s,a)
-\alpha \sum_{s,a} p(s,a) \log p(s,a) 
\nonumber
\\
&&
-(\beta - \alpha) \sum_{s,a} p(s,a) \log \left(\sum_b p(s,b) \right) 
.
\label{eq_average_reward_p2}
\eea
\noindent
We note that $R_p$ is continuous in $p$ in the whole domain and differentiable for $0<p(s,a)<1$. Therefore, concavity of $R_p$ is equivalent to show that the second order derivative in any direction from any $0<p(s,a)<1$ for all $(s,a)$ (i.e., away from the simplex boundaries) are non-positive. From any such $p(s,a)$, consider the direction $u(s,a)$ (we only need to consider directions such that $\sum_{s,a} u(s,a) =0$ so that $p(s,a)+\eta u(s,a)$ is a probability distribution for small enough values of $\eta$). To compute the 2nd-order directional derivatives, we first write the average reward by moving $p(s,a)$ in the direction $u(s,a)$ by an amount $\eta$
\bea
R_{p,u}(\eta)
 &=&
 \sum_{s,a} (p(s,a) + \eta u(s,a)) r(s,a)
 \nonumber
 \\
 &&-\alpha \sum_{s,a} (p(s,a) + \eta u(s,a)) 
\log (p(s,a) + \eta u(s,a))
 \nonumber
 \\
 &&-(\beta - \alpha) \sum_{s,a} (p(s,a) + \eta u(s,a)) \log \left(\sum_b (p(s,b) + \eta u(s,b)) \right) 
 .
 \label{eq_average_reward_p3}
\eea
\noindent
From here, the 2nd-order directional derivative is
\begin{align}
    R^{''}_{p,u}(\eta) \Bigr\rvert_{\eta=0}  =&
    -\alpha 
\sum_s \left( \sum_{a} \frac{u^2(s,a)}{p(s,a)}  - \frac{ (\sum_a u(s,a)  )^2 }{\sum_a p(s,a)}\right) 
-\beta \sum_s \frac{ (\sum_a u(s,a)  )^2 }{\sum_a p(s,a)}\leq 0
\label{eq_2nd_dir_deriv}
\end{align}
\noindent
for $\alpha \geq 0$ and $\beta \geq 0$.
To see this, note that the second term in the right-hand side is non-positive for all directions iff $\beta \geq 0$ (it is zero in the trivial direction $u(s,a)=0$ for all $s$ and $a$, but also in the action directions such that $\sum_a u(s,a)=0$ for all $s$). 
Likewise, the first term is non-positive for al directions iff $\alpha \geq 0$; it is only zero for the trivial direction and the parallel direction $u(s,a) \propto p(s,a)$. This is because each term in the sum over $s$ is non-negative, as it can be seen by using Schwartz's inequality $(\sum_a x_a^2) (\sum_a y_a^2) \geq (\sum_a x_a y_a )^2$ 
with $x_a = u(s,a)/\sqrt{p(s,a)}$ and $y_a = \sqrt{p(s,a)}$.  
As the parallel direction is excluded because it does not obey the normalization condition $\sum_{s,a} u(s,a) = 0$, then the first term is strictly negative iff $\alpha > 0$. 
In summary, the 2nd-order derivative in any direction from any $p(s,a)$ not in the simplex boundaries is non-positive for $(\alpha \geq 0, \beta \geq 0)$, which implies that the average total reward $R_p$ is concave everywhere due to continuity, concluding the proof.

Along with the fact that the constraints (\ref{eq_cond1}-\ref{eq_cond3}) are linear in $p(s,a)$, and thus define a convex set, the optimization problem in Eqs. (\ref{eq_average_reward_max_p}-\ref{eq_cond3}) is convex.


Previous work has shown that the average total reward for the cases $(\alpha,\beta)=(1,1)$ \cite{peters2010relative}, $(\alpha,\beta)=(1,0)$ \cite{neu2017unified} and $(\alpha,\beta)=(0,1)$ \cite{hazan2019provably} is concave in $p(s,a)$.
Thus, Theorem \ref{th_concave_1} can also be readily obtained by linearly combining the second and third cases with non-negative coefficients $(\alpha \geq 0, \beta \geq 0)$. 
The novelty of our proof is to show that Eq. (\ref{eq_average_reward_p2}) is concave {\em iff} $(\alpha \geq 0, \beta \geq 0)$ if we allow any direction, including the parallel one.

\subsection{Critical points}

We now find the critical points of the average total reward (\ref{eq_average_reward_p}) as a function of $p$ under the constraints (\ref{eq_cond1}-\ref{eq_cond3}) using Lagrange multipliers. The Lagrangian $L \equiv L(p,V,\lambda)$ is
\bea
L
&=&
\sum_{s,a} p(s,a) r(s,a)
-\alpha \sum_{s,a} p(s,a) \log p(s,a) 
-(\beta - \alpha) \sum_{s,a} p(s,a) \log \left(\sum_b p(s,b) \right) 
\nonumber
\\
&&+ 
\sum_{s'} V(s') \left(  \sum_{s,a} p(s'|s,a) p(s,a) - \sum_b p(s',b)  \right) + 
\lambda \left( \sum_{s,a} p(s,a) - 1 \right)
\;,
\label{eq_L}
\eea
\noindent
where the $V$ and $\lambda$ are multipliers, and condition (\ref{eq_cond1}) will be shown below to be automatically satisfied by the optimal solution. Differentiating with respect to each $p(s,a)$ leads to the equation for the critical points of the average total reward
\begin{align}
\frac{\partial L}{\partial p(s,a)}
&=
r(s,a) -\alpha -\alpha \log p(s,a)-(\beta - \alpha) \log \left(\sum_b p(s,b) \right) \nonumber
\\ 
&-
(\beta - \alpha) \frac{\sum_a p(s,a)}{\sum_b p(s,b)}  
+\sum_{s'} V(s') p(s'|s,a) -V(s)
+ \lambda \nonumber \\&= 0 
\;.
\label{eq_L_diff}
\end{align}
\noindent
After simplification and solving for $p(s,a)$, we obtain for $\alpha \neq 0$ the equation
\be
 p(s,a) = e^{(\lambda - \beta)/\alpha}  
 \left( \sum_b p(s,b) \right)^{1-\beta/\alpha} e^{A(s,a)/\alpha}
 \;, 
 \label{eq:p1}
\ee
\noindent
where we have defined the {\em advantage} function
\be
  A_V(s,a) = r(s,a) + 
  			\sum_{s'} V(s') p(s'|s,a)
  			-V(s)
  \;.
  \label{eq_advantage}
\ee
\noindent
Note that $V(s)$ can be interpreted as the value of being at state $s$, and thus states with larger $V(s)$ are preferred over the others \cite{peters2010relative,puterman2014markov}.

Summing over $a$ in Eq. (\ref{eq:p1}) and solving for $\sum_a p(s,a)$, we obtain for $\beta \neq 0$
\be
\sum_a p(s,a) = e^{\lambda/\beta-1} 
   \left( \sum_a e^{A_V(s,a)/\alpha}   \right)^{\alpha/\beta}
\ee
\noindent
Inserting this equation into the right-hand side of Eq. (\ref{eq:p1})
\be
p(s,a) = e^{\lambda/\beta -1 }  
\left( \sum_b e^{A_V(s,b)/\alpha} \right)^{\alpha/\beta-1} e^{A_V(s,a)/\alpha}
\;.
\label{eq:p2}
\ee
\noindent
Normalization of $p(s,a)$ implies that
\be
p(s,a) = \frac{1}{Z_V}  
\left( \sum_b e^{A_V(s,b)/\alpha} \right)^{\alpha/\beta-1} e^{A_V(s,a)/\alpha}
\;, 
\label{eq:p3}
\ee
\noindent
with normalization constant
\be
Z_V = \sum_s \left( \sum_a e^{A_V(s,a)/\alpha} \right)^{\alpha/\beta}
\;, 
\label{eq:Z}
\ee
\noindent
and the multiplier $\lambda$ is given by
\be
\lambda = \beta (1- \log Z_V)
\;.
\label{eq:lambda}
\ee
\noindent
Eqs. (\ref{eq:p3}-\ref{eq:Z}) formally define the optimal state-action joint probability $p^{*}(s,a)$ as a function of the optimal multipliers $V^*$. It is clear that for any value of the $V$ the probabilities are non-negative, and therefore condition (\ref{eq_cond1}) is automatically satisfied. 
The normalization condition (\ref{eq_cond2}) is obviously satisfied. It is shown below that condition (\ref{eq_cond3}) is satisfied by the optimal values $V^*$, thus fully determining $p^{*}(s,a)$. 

Once the optimal $V^*$ is obtained, the optimal policy is 
\be
 \pi^*(a|s) = \frac{e^{A_{V^*}(s,a)/\alpha}}
 	{\sum_b e^{A_{V^*}(s,b)/\alpha}} 
 \;,
\label{eq:pi_opt}
\ee
\noindent
the optimal joint action-state probability is
\be
p^*(s,a) = \frac{1}{Z_{V^*}}  
\left( \sum_b e^{A_{V^*}(s,b)/\alpha} \right)^{\alpha/\beta-1} e^{A^*(s,a)/\alpha}
\label{eq:p_s_a_opt}
\ee
\noindent
and the stationary state distribution becomes
\be
  p^*(s) = \frac{1}{Z_{V^*}}  
  \left( \sum_a e^{A_{V^*}(s,a)/\alpha} \right)^{\alpha/\beta}
  \;.
\label{eq:p_s_opt}
\ee
\noindent

\subsection{Dual function}

We consider the primal problem the one defined in (\ref{eq_average_reward_max_p}-\ref{eq_cond3}). The dual problem consists in minimizing the dual function of the Lagrangian with respect to the multipliers.
The dual function for $\alpha>0$ and $\beta>0$ is obtained by replacing $p(s,a)$ in the (log terms of the) Lagrangian (\ref{eq_L}) by their critical values in Eq. (\ref{eq:p2}) as a function of the multipliers. 
This substitution leads to the dual function (Appendix \ref{derivation})  
\be
L^{d}(V,\lambda) = \beta \log Z_V
\label{eq_L_dual}
\;.
\ee
\noindent
Due to duality, maximizing the average total reward in the primal problem (\ref{eq_average_reward_max_p}-\ref{eq_cond3}) is equivalent 
to minimizing the partition function $Z_V$ in Eq. (\ref{eq:Z}) with respect to the $V$ {\em without} any constraint (see Appendix \ref{derivation} to confirm that $\lambda$ can be chosen to obey the normalization constraint (\ref{eq:p2_norm})).
Therefore, the initial constrained concave optimization problem has been transformed into an unconstrained convex optimization one where $\log Z_V$ is to be minimized with no constraints over the $V$, and thus 
\be
  V^* = \argmin_V \log Z_V
  \label{eq:V_opt}
  \;.
\ee 
\noindent
From here, the optimal policy and state probability are found using Eqs. (\ref{eq:pi_opt}-\ref{eq:p_s_opt}).

Although from duality it is clear that $\log Z_V$ is convex in $V$, this fact can also be directly checked, as follows. This exercise will also show under what conditions the $\log Z_V$ is strictly convex, which will be important to show below uniqueness of the optimal solution. 
We first write the dual function as 
\be
L^{d}(V) = \beta \log Z_V = \beta \log \left( \sum_s 
  e^
  {\frac{\alpha}{\beta} \log( \sum_a e^{A_V(s,a)/\alpha} ) }  \right)
  \label{eq:Z_log_log}
\ee   
\noindent
and notice that $A_V(s,a)$ is convex in the $V$ (indeed, it is linear).
Then, by noticing that $\log Z_V$ is an increasing function of $A_V$, we just need to show that it is convex in $A_V$, as the composition of an increasing convex function with a convex function is convex (see Appendix \ref{convexity_cond}). 
Further, we note that Eq. (\ref{eq:Z_log_log}) is the composition of two identical functions of the log-sum-exp form $h(x)=\log( \sum_i e^{cx_i})$ with (possibly different) positive $c$. Both are increasing, and therefore according to the previous composition rule for convexity it remains to be seen that $h(x)$ is convex to show that $\log Z_V$ is convex in $V$.

Although convexity of $h(x)$ is a well-known fact \cite{boyd2004convex}, here we explicitly prove it to get additional information that will be relevant to show uniqueness of the optimal solution. Let us calculate the 2nd-order directional derivative of $h(x)$ in a direction $y$. For that, consider $h(x + \eta y)$. Taking the 2nd-order derivative with respect to $\eta$ and noticing that the we can take $c=1$ without affecting the conclusions on convexity, we find 
\be
\frac{\partial^2 h(x) }{ \partial \eta ^2} |_{\eta=0}=
\frac{(\sum_i y^2_i e^{x_i})(\sum_i e^{x_i})-(\sum_i y_i e^{x_i})^2}{(\sum_i e^{x_i})^2}
\geq 0 
\;,
\label{eq:h_second_order}
\ee
\noindent
where in the inequality we have used the fact that the denominator is positive, as the action set in non-empty for all $s$, and that the numerator is non-negative in virtue of the Schwartz's inequality $(\sum_i w_i \tilde x_i^2) (\sum_i w_i \tilde y_i^2) \geq (\sum_i w_i \tilde x_i \tilde y_i )^2$ ($w_i >0 $) with $w_i = e^{x_i} $, $\tilde x_i = 1$ and $\tilde y_i = y_i$.  
This shows that for any $x$ and for any direction $y$ the 2nd-order directional derivative is non-negative, showing that $h(x)$ is convex, which finally implies that $\log Z_V$ is convex in $A_V$ being a composition of $h(x)$ with itself (with possibly different positive values of $c$).

Now, looking at Eq. (\ref{eq:h_second_order}) and using Schwartz's inequality, it is clear that $h(x)$ is strictly convex in all directions $y$ except in the direction $y_i = 1$. As $\log Z_V$ is the composition of the increasing convex $h(x)$ with itself, this implies that $\log Z_V$ is strictly convex in all directions except in the diagonal direction where $A_V(s,a)$ changes by a constant $A_0$ for all $(s,a)$ (that is, $A(s,a) \rightarrow A(s,a) + A_0$). In the diagonal direction, $\beta \log Z_V$ is linear and increasing exactly as $A_0$, as it can be easily checked.   

Also, from duality the location $V^*$ of the minimum of the dual function, Eq. (\ref{eq:V_opt}), makes the optimal joint state-action probability in Eq. (\ref{eq:p_s_a_opt}) to satisfy condition (\ref{eq_cond3}). In other words, while conditions (\ref{eq_cond1}-\ref{eq_cond2}) are automatically fulfilled by the state-action probability Eq. (\ref{eq:p_s_a_opt}) for any choice of the $V$, condition (\ref{eq_cond3}) is only fulfilled when $V=V^*$ (here equality is understood as matching a member of the class of critical points). 
To see that for $V=V^*$ condition (\ref{eq_cond3}) is satisfied, let us compute the critical points of the dual function by taking derivatives with respect the $V$ in Eq. (\ref{eq:Z_log_log}),
\bea
\frac{\partial L^{d}(V) }{\partial V(s)} &=&
\frac{\beta}{Z_V} 
\Big(   
-\beta^{-1}
e^
{\frac{\alpha}{\beta} \log( \sum_a e^{  A_V(s,a) /\alpha} ) }
\nonumber
\\
&&+
\beta^{-1} \sum_{s'}
e^
{\frac{\alpha}{\beta} \log( \sum_a e^{  A_V(s',a) /\alpha} ) }
\frac{\sum_a p(s|s',a) e^{  A_V(s',a) /\alpha} }
{\sum_a e^{  A_V(s',a) /\alpha} }
\Big) \nonumber
\\&&=
0
\label{eq:critical_points_dual}
\;.
\eea 
\noindent
Rearranging terms, this implies that
\bea
&& 
\frac{1}{Z_V}
e^
{\frac{\alpha}{\beta} \log( \sum_a e^{  A_V(s,a) /\alpha} ) }
= 
\sum_{s',a} p(s|s',a) 
\frac{
	e^
	{\frac{\alpha}{\beta} \log( \sum_a e^{  A_V(s',a) /\alpha} ) }
	e^{  A_V(s',a) /\alpha} }
{Z_V \sum_b e^{  A_V(s',b) /\alpha} }
\;.
\label{eq:self_consistency}
\eea

\noindent
Now, using Eq. (\ref{eq:p3}), the left-hand side equals $p(s) = \sum_a p(s,a)$ and the right-hand side equals $\sum_{s',a} p(s|s',a) p(s',a)$. Thus at the critical point(s) Eq. (\ref{eq:self_consistency}) is satisfied, and therefore condition (\ref{eq_cond3}) holds when $V=V^*$. 
Obviously from the above, if $V \neq V^*$, then condition (\ref{eq_cond3}) does not hold. In summary, the optimal solution is found at the critical points of the dual function (which can be at $V(s) = \pm \infty$ for some $s$).  

The above results lead to

\begin{theorem}[]
	The optimal $p^*(s,a)$ maximizing the average total reward $R_p$ under the constraints (\ref{eq_cond1}-\ref{eq_cond3}) for $\alpha>0$ and $\beta>0$ is unique and satisfies 
	\bea
	&& p^*(s,a) =
            \frac{1}{Z_{V^*}}  
		e^
	{(\frac{\alpha}{\beta}-1) \log( \sum_b e^{A_{V^*}(s,b)/\alpha} ) 
	+ A_{V^*}(s,a)/\alpha}
	\label{eq:p3_th}
	\eea
	\noindent
	with
	\be
	Z_{V^*} = \sum_s 
	e^
	{\frac{\alpha}{\beta} \log( \sum_a e^{A_{V^*}(s,a)/\alpha} ) }
	\;, 
	\label{eq:Z_th}
	\ee	
	\noindent
	and
	\be
	A_{V^*}(s,a) = r(s,a) + 
	\sum_{s'} V^*(s') p(s'|s,a)
	-V^*(s)
	\;,
	\label{eq:A_V*}
	\ee
	\noindent
	where the $V^*$ minimize the (convex) dual function
	\be
	L^{d}(V) = \beta \log Z_V = \beta \log \left( \sum_s 
	e^
	{\frac{\alpha}{\beta} \log( \sum_a e^{A_V(s,a)/\alpha} ) }  \right)
	\;,
	\label{eq:L_dual_final}
	\ee
	\noindent
	that is,
	\be
	V^* = \argmin_V \log Z_V
	\;.
	\ee
	\noindent
	The maximum average total reward is 
	\be
	  R_{max} = \beta \log Z_{V^*}
	  \;.
	  \label{eq:R_max_opt_final}
	\ee
        For $s$ such that $p(s)>0$, the optimal policy is uniquely defined by 
        \be
        \pi^*(a|s) = \frac{e^{A_{V^*}(s,a)/\alpha}}
	{\sum_b e^{A_{V^*}(s,b)/\alpha}} 
	\;.
 	\label{eq:pi_opt_final}
        \ee
        \noindent 
        while for $s$ such that $p(s)=0$ the policy can be arbitrarily defined. 
	\label{th_dual}
\end{theorem}
\noindent
It remains to show uniqueness and the validity of Eq. (\ref{eq:R_max_opt_final}).

That $A^*(s,a)$ and thus $p^*(s,a)$ are unique can be seen from the strict convexity of $\log Z_V$ as a function of the $A(s,a)$ in directions away from the diagonal direction. First, assume that there are two optimal (minima) solutions $V^*$ of $\log Z_V$ such that $A^*_1(s,a)$ and $A^*_2(s,a)$ are different for some $s$ and $a$. Then, it is not possible that $A^*_1 = A^*_2 + d$ with a constant $d \neq 0$; otherwise, the value of $\log Z_V$ would be different for the two optimal solutions as $\beta \log Z_V$ is linear in $d$ with slope $1$, contradicting that both are optimal minima. 
Therefore, the line that joins $A_1^*$ and $A_2^*$ is not in the diagonal direction. But we already know that $\log Z_V$ in any non-diagonal direction is strictly convex in $A(s,a)$. Therefore, there should a lower value of $\log Z_V$ in between the points $A_1^*$ and $A_2^*$, contradicting that they were the absolute minima. 
This concludes the proof that $A^*(s,a)$ and thus $p^*(s,a)$ are unique.

The previous reasoning also reveals that there are degeneracies in the values of $V^*$, that is, their optimal values are not unique: from Eq. (\ref{eq:A_V*}) if $V^*(s)$ is an optimal solution, then $V^*(s) + d$ for any $d$ is also an optimal solution, as the value of $A^*$ remains unchanged. 
Degeneracies of $V^*$ are not restricted to the diagonal direction. For instance, if $p(s'=s|s,a)=1$ for some $s$ and for all actions $a$ in that state, then any value of $V^*(s)$ is valid --the value function for the "isolated", self-connected, node $s$ 
does not affect the rest of $V^*$. 
Note that this node would be unreachable from other nodes, but it has generally assigned an non-zero optimal probability, $p^*(s,a) > 0$, to increase state entropy.

To conclude the proof of the theorem, Eq. (\ref{eq:R_max_opt_final}) results from strong duality, or more directly from substitution of the optimal state-action probability Eq. (\ref{eq:p3_th}) into Eq. (\ref{eq_average_reward_p2}).

\subsection{Solvable toy example}
\label{Sec:toy_example}

\begin{figure}[t]
    \centering
    \includegraphics[width=\columnwidth]{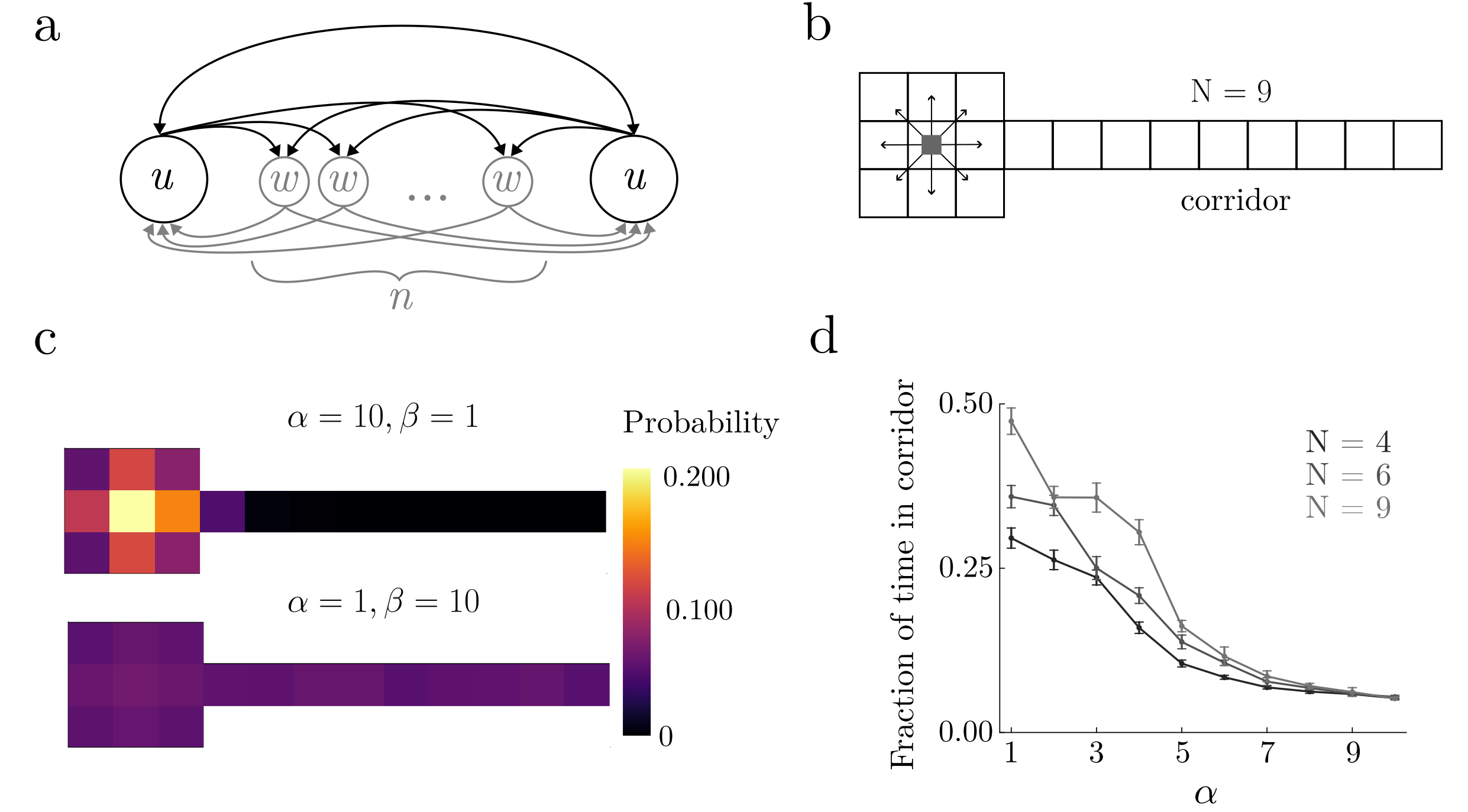}
    \caption{Action and state entropy regularizers have different effects on optimal policies. (a) Schematic of a toy example. There are two ``outer" states from which it is possible to transition to any other state ($n+2$ actions), and $n$ ``inner" states, from which there are only transitions to the outer states. (b) Grid world arena. The agent (grey square) has nine available actions unless constrained by the walls. Therefore, in the middle of the corridor, there are only three actions: \texttt{left, right, nothing}. (c) Heatmap of state distributions for two agents with the associated hyperparameters. An action-focused agent ($\alpha > \beta$, top) prefers the middle of the room, given the action availability, wheres a state-focused agent ($\beta > \alpha$, bottom) prefers being spread out evenly in state space. (d) Fraction of time spent in the corridor for various $\alpha$ and $\beta$ values, where $\alpha\beta = 10$, for various corridor lengths.}
    \label{fig:schematic}
\end{figure}

We consider here a simple MDP, described in Fig. \ref{fig:schematic}a with $r(s,a)=0$ to study how states and actions are occupied for different values of $\alpha$ and $\beta$ without the effect of rewards. 
This example is designed such that there are few states with many possible actions and many states with few actions, so that the effects of manipulating the weights of the action and state entropies are the clearest.  
There are two "outer" states of the same kind, and $n>0$ "inner" states of the same kind. From an outer state, the agent can take the action of going to the other outer state or go to any of the inner states. From the inner state, it is only possible to choose the action of going to one outer state or to the other. Transitions are deterministic given the action.
Thus, there are a total of $n+2$ states; from the outer states there are $n+1$ possible actions and from the inner states there are $2$ possible actions.  
We expect that the state distribution for an agent with $\alpha \gg \beta$, which favors action entropy over state entropy, would show a larger probability for the outer than for the inner states.
The reverse is expected for an agent with $\alpha \ll \beta$.
Because of the symmetry of the MDP and the strict convexity of the dual function in any non-diagonal direction, the optimal values should be identical for the two outer states, and also identical for the $n$ inner states, but they can differ between them. Let us call these values $V(s_{outer})=u$ and $V(s_{inner})=w$, and choose $w=0$ arbitrarily because only relative values are important.
The dual function takes the form 
\bea
	L^d(u) &=&
	\beta \log Z_{u} \nonumber
        \\
        &=&  
	\beta \log 
	\left(
		2(1+n e^{-u/\alpha})^{\alpha/\beta} + 2^{\alpha/\beta} n e^{u/ \beta}
	\right)
 \nonumber
\eea
\noindent
and its critical point $dL^d(u)/du=0$ is given by
\be
(1 + n e^{-u/\alpha})^{\alpha/\beta-1} e^{-u/\alpha}
= 
2^{\alpha/\beta-1} e^{u/\beta}
\;.
\label{eq:derivative_toy}
\ee
Now we consider three relevant cases:

$\bullet$ $(\alpha=1,\beta=1)$. From Eq. (\ref{eq:derivative_toy}) we have $u=0$. Therefore, the optimal values are identical for all nodes. However, the probabilities of outer and inner states are not the same. From Eq. (\ref{eq:pi_opt}) it is clear that the optimal policy is uniform in all the actions, that is, the probability of going to an inner or outer state from an outer state is the same, and equals $1/(n+1)$. Using condition (\ref{eq_cond3}), this leads to $p(s_{outer})=(n+1)/(2(2n+1))$ and $p(s_{inner})=1/(2n+1)$, and therefore outer are more likely than inner states. This makes sense, as outer states can achieve higher action entropy and a balance between action and state entropies is sought by the agent.

$\bullet$ $(\alpha=1,\beta=0)$. Taking the limit $\beta \rightarrow 0$ in Eq. (\ref{eq:derivative_toy}), we obtain $u=\log ( 1+ \sqrt{1+8n} ) - 2 \log 2 \geq 0$. Therefore, the value of the outer states grows with $n$ compared to the value of the inner states. 
In this case, the transition probability from an outer to the other outer state is $\pi_n=1/(1+ne^{-u})$, which is $\geq$ than the uniform $1/(n+1)$. The transition probability from an outer to a given inner state is $1/(n+e^u) \leq 1/(n+1)$. 
As a result, $p(s_{outer})=1/(2(2-\pi_n))$, larger than in the previous example. 

$\bullet$ $(\alpha=0,\beta=1)$. Taking the limit $\alpha \rightarrow 0$ in Eq. (\ref{eq:derivative_toy}), we obtain $u=-\log(n/2) / 2$ for $n>1$ and $u=0$ for $n=1$. Therefore, the value of the outer states decreases with $n$. In this case $\pi_n=1/(1+ne^{-u})$ is smaller than the uniform probability $1/(n+1)$ for $n>2$, and $p(s_{outer})=1/(2(2-\pi_n))$ is smaller than for the first case $(\alpha=1,\beta=1)$.

\subsection{Limit cases}

The cases $(\alpha>0,\beta=0)$ and $(\alpha=0,\beta>0)$ can be considered as the limits of $\beta \rightarrow 0^+$ and $\alpha \rightarrow 0^+$ respectively. We now show in turn that the limits are well-defined and that for the first case it corresponds to the actual known solution for $(\alpha,\beta=0)$.

$\bullet$ $(\alpha>0,\beta \rightarrow 0^+)$. We rewrite the equation of the critical points, Eq. (\ref{eq:critical_points_dual}), as
\begin{align}
&	\sum_a e^{A_V(s,a)/\alpha}
	= 
	\sum_{s'}
	\frac{(\sum_a  e^{A_V(s',a) / \alpha}  )^{\alpha / \beta -1} }{ (\sum_a  e^{A_V(s,a) / \alpha}  )^{\alpha / \beta -1} }
		\sum_a e^{A_V(s',a)/\alpha} p(s|s',a)
		\;.
 \label{eq:L_V_der}
\end{align}
\noindent
for all $s$. We restrict ourselves to a MDP where it is possible to choose actions such that there is a positive probability of reaching any state from any other state in a finite number of steps. 
Define $e^{\eta_V(s)/\alpha} = \sum_a e^{A_V(s,a)/\alpha}$, and assume that in the limit $\beta \rightarrow 0$ the $\eta_V(s)$ are not identical for all $s$ and that all of them are finite. 
This implies that there exists one state $s'$ with a successor state $s$ (that is, a state accessible from $s'$ using some action $a$: $p(s|s',a)>0$) for which $\eta_V(s')>\eta_V(s)$.
Looking at Eq. (\ref{eq:L_V_der}), it is clear that then $\eta_V(s) = \infty$, as at least one of the ratios in the right-hand side of the equation diverges as $\beta \rightarrow 0$ and the rightmost sum over actions is finite and positive.  
This contradicts the initial assumption, and 
therefore $\eta_V(s) = \alpha \log ( \sum_a e^{A_V(s,a)/\alpha} ) = \eta_V$ are identical for all $s$.
If $\eta_V$ is infinite, then the dual function will take infinite value, which cannot correspond to its minimum (e.g., choosing $V(s)=0$ for all $s$ leads to a lower value of the dual function for finite $r(s,a)$).
In conclusion, all $ \alpha \log ( \sum_a e^{A_V(s,a)/\alpha} ) = \eta_V$ must be finite and independent of $s$, which implies, using Eq. (\ref{eq:L_V_der}), that
\be
\sum_a e^{A_V(s,a)/\alpha}
=
\sum_{s',a}
e^{A_V(s',a)/\alpha} p(s|s',a)
\;.
\ee
\noindent
Therefore, in this limit, the dual function becomes $L^d(V) = \beta \log ( \sum_s e^{\eta_V/\beta}) = \eta_V$.
As the dual function ought to be minimized with respect to $V$,  and thus with respect to $\eta_V$, then we have proved 

\begin{theorem}[]
	The optimal $p(s,a)$ maximizing the average total reward $R_p$ under the constraints (\ref{eq_cond1}-\ref{eq_cond3}) for $\alpha>0$ and $\beta \rightarrow 0^+$ is unique and satisfies
	\be
	p^*(s,a) = \frac{e^{A_{V^*}(s,a)/\alpha}}
	{\sum_{s,b} e^{A_{V^*}(s,b)/\alpha}}  
	\label{eq:p_sa_th_beta0}
	\ee		
	\noindent
	where $V^*$ is such that the constraint 
	\be
		\eta_{V^*} = \alpha \log \left( \sum_a  e^{A_{V^*}(s,a) / \alpha} \right) 
		\label{eq:constraint_eta_V}
	\ee
	\noindent
	is satisfied for all $s$, and $\eta_{V^*}$ is chosen such that it is the smallest possible real number. 
		
	The maximum average total reward is
	\be
	R_{max} = \eta_{V^*}
	\;.
	\label{eq:R_max_opt}
	\ee
	\label{th_beta_lim0}
\end{theorem}
\noindent
The equations result from direct substitution of those in Theorem \ref{th_dual} and taking the appropriate limit. We remark that the optimal solution is identical to the exact case $(\alpha,\beta=0)$ \cite{neu2017unified}, so here we have shown that the case $(\alpha,\beta=0)$ corresponds to the limit of $(\alpha,\beta)$ as $\beta \rightarrow 0$.
We note that the constraint (\ref{eq:constraint_eta_V}) is satisfied in the example of Sec. (\ref{Sec:toy_example}).

$\bullet$ $(\alpha \rightarrow 0^+,\beta>0)$.
Taking the limit $\alpha \rightarrow 0$ in Eq. (\ref{eq:L_dual_final}) leads to a novel form of the dual function,
\be
 L^d(V) = \beta \log \left( \sum_s e^{\max_a A_V(s,a) / \beta } \right)
 \;,
 \label{eq:L_d_alpha0}
\ee
\noindent
which should be minimized as a function of $V$. 
It can be checked that the Eq. (\ref{eq:L_d_alpha0}) holds in the example of Sec. (\ref{Sec:toy_example}). 
The form of the dual function makes sense given that the optimal policy in Eq. (\ref{eq:pi_opt_final}) becomes deterministic in the limit, except for the possible degeneracy of actions having the same $\max_a A_{V^*}(s,a)$. Therefore, in general the optimal policy is not unique.
The optimal policy can induce a periodic Markov process, in which case $p(s,a)$ might be understood as the probability of finding the process in $(s,a)$ when observed at a randomly chosen time. 
For instance, consider the example consisting of a circular chain with $n$ states where transitions are allowed to any of the two neighbors and $r(s,a)$ is uniform: the policy that moves the agent from one state to its right neighbor state induces an uniform probability in the chain, maximizing state entropy, but the process is periodic.   
Importantly, if we take in this example $\alpha$ to be very small but non-zero, then the problem is regularized and symmetries are broken, in the sense that the optimal policy is unique and it corresponds to the uniform action distribution (due to Theorem \ref{th_dual}), which moves the agent to any of the two neighboring states with equal probability.  

$\bullet$ $(\alpha \rightarrow 0^+,\beta \rightarrow 0^+)$.
Taking the limit of $\alpha \rightarrow 0^+$ in Eq. (\ref{eq:constraint_eta_V}) leads to the constraint
	\be
\eta_{V^*} = \max_a  A_{V^*}(s,a)  
\label{eq:constraint_eta_V_alphabeta0}
\ee
\noindent
for all $s$, where $V^*$ are chosen such that $\eta_{V^*}$ is minimized. Putting the solution in the more recognizable format
\be
  V^*(s) = \max_a \left( r(s,a) -\eta_{V^*} + 
  	\sum_{s'} p(s'|s,a) V^*(s')  \right)
  	\;,
\ee
\noindent
it becomes apparent that it corresponds to the standard solution for the case where the total reward is simply $R_{\pi}(s,a)=r(s,a)$ \cite{puterman2014markov}.
Therefore, one recovers the well-known average-reward Bellman optimality equation for the RL problem with no state-action entropy, whose policy is deterministic except for possible action ties.  

\subsection{Experiments}

We simulated an agent for different $(\alpha,\beta)$ in an arena with a room and a narrow corridor (Fig. \ref{fig:schematic}b; Appendix \ref{cat-room}). In the room there are more actions than in the corridor. For $\alpha \gg \beta$ the agent seeks action over state entropy, and therefore the room is occupied while the corridor is left empty (Fig. \ref{fig:schematic}c, top). In contrast, for $\beta \gg \alpha$ the agent seeks state over action entropy, and therefore it occupies both the room and corridor more uniformly (bottom). The effect of $\alpha$ on the fraction of time that the agent spends in the corridor is smooth and decreases with it for various corridor lengths (Fig. \ref{fig:schematic}d).

Rather than using gradient descent over the $V$ directly on the dual function in Eq. (\ref{eq:L_dual_final}) to find the optimal $p^*(s,a)$, we developed an iterative scheme that empirically is shown to effectively find the optimal solution (see Appendix \ref{iterations}).

\section{Discussion}

Optimal control and reinforcement learning approaches deal with the problem of solving or obtaining optimal courses of action for a given measure. While it has been postulated that a scalar reward signal is a sufficient measure to obtain practically any desirable behavior \cite{sutton1998introduction,silver2021reward}, this approach has severe limitations that have been highlighted throughout the years \cite{mcnamara1986common,schmidhuber1991possibility,klyubin2005empowerment,singh2009rewards}. Here, we have shown that seemingly disparate entropy augmentations to the scalar reward signal are in fact captured by a general framework whose unique solutions we have found. Our central contribution is to provide a general dual function formalism (convex and constraint-free) for arbitrary mixtures of action and state entropy reward regularizers. Standard examples with only action entropy, only state entropy, or balanced action-state entropy are particular cases of our formalism (see Introduction). Other cases with no entropy correspond to certain limits. KL approaches simply correspond to a redefinition of the reward function in the general maximum mixed action-state entropy formalism. 

We surmise that general mixed action-state entropy regularization is important because, 
by changing the mixture hyperparameters $\alpha$ and $\beta$ independently, the agent can be pushed to learn different aspects of an MDP, namely, the action availability and the state availability, respectively.
Previous work has shown that KL action entropy regularization ($\alpha=1,\beta=0$ with the default policy being the previously learnt policy) is critical to speed up and stabilize learning \cite{schulman2015trust,neu2017unified,haarnoja2018soft}. 
By allowing arbitrary mixtures $(\alpha,\beta)$ of KL action and state regularization with a default policy and state distributions corresponding to the previously learnt ones, we propose that one could more effectively control the learning process by: (1) biasing the agent to be conservative regarding the policy update by using KL action regularization, or (2) biasing it to be more conservative about the state distribution updates by using KL state regularization. In the first case, the agent will try to repeat behaviors previously learnt, but could fall into unexpected, potentially dangerous, states; while in the second case, the agent will try to repeat visiting previously visited states, but could end up performing undesired actions. 
Our framework allows adjusting each depending on the desired learning process.
A complementary view of the same problem is also relevant: in the first case, the agent is biased to explore state space more, while in the second case, it prefers exploring action space more. We conjecture that balancing action and state entropy is critical to optimize exploration and learning.

\section*{Acknowledgments}
This work is supported by the Howard Hughes Medical Institute (HHMI, ref 55008742) and MINECO (Spain; BFU2017-85936-P) to R.M.-B, and MINECO/ESF (Spain; PRE2018-084757) to J.R.-R.

\bibliographystyle{unsrtnat}
\bibliography{undiscounted}


\newpage
\appendix
\onecolumn

\section{Convexity conditions}
\label{convexity_cond}

As it is well known, if $h(x)$ is increasing and convex in $x \in \mathbb{R}^n$ and $g_i(y)$, $i=1,...,n$ are convex in $y$, then for $y=\eta y_a + (1-\eta) y_b$ with $0\leq \eta \leq 1$
we have $h(g_1(y),...,g_n(y)) 
\leq h(\eta g_1(y_a) + (1-\eta) g_1(y_b) ,..., \eta g_n(y_a) + (1-\eta) g_n(y_b)) 
\leq  \eta h(g_1(y_a),...,g_n(y_a) + (1-\eta) h(g_1(y_b),...,g_n(y_b) )$, where in the first inequality we have used convexity of the $g_i$, that is, $g_i(\eta y_a + (1-\eta) y_b) \leq  \eta g_i(y_a) + (1-\eta) g_i(y_b) $, and that $h$ is increasing, and in the second inequality we have used convexity of $h$.

\section{Derivation}
\label{derivation}

Here we derive Eq. (\ref{eq_L_dual}).
Starting from the definition of the Lagrangian (\ref{eq_L}), we replace the log terms by the critical values of $p(s,a)$ in Eq. (\ref{eq:p2}), which leads to   
\bea
L^{d}(V,\lambda)
&=&
\sum_{s,a} p(s,a) \left\{
r(s,a)
-\alpha \left( \lambda/\beta - 1 + (\alpha/\beta -1) \log (\sum_b e^{A_V(s,b)/\alpha})  + A_V(s,a)/\alpha \right)
\right.
\left.
\label{eq_L_append}
\right. \nonumber
\\
&& 
\left.
-(\beta - \alpha) \left(\lambda/\beta - 1 
+ \alpha/\beta \log (\sum_b e^{A_V(s,b)/\alpha}) \right)
\right\} \nonumber
\\
&&+ 
\sum_{s'} V(s') \left(  \sum_{s,a} p(s'|s,a) p(s,a) - \sum_b p(s',b)  \right) 
\\
&& = \beta \log Z_V
\nonumber
\;,
\eea
\noindent
where we have used the normalization constraint $\sum_{s,a} p(s,a)=1$ in the first equality, and we have used the definition of $A_V(s,a)$, Eq. (\ref{eq_advantage}), in the second one.

\section{Simulation details}

\subsection{Grid world} \label{cat-room}

\paragraph{Environment} The arena is composed of one room having size $3 \times 3$ cells and a "corridor" of size $1 \times N$ cells, which is connected to the room at the middle of its right side (see Fig. \ref{fig:schematic}b). 

\paragraph{States, Actions and Transitions} The (discrete) state is the position of the cell where the agent stays. The agent has up to 9 actions: to move to one of the available neighbor cells around the current location (including diagonal motions), or to stay on the current position. The position to which it moved becomes its state at the following time step.  Moving through borders 
is not possible, and thus those actions are not available in the boundary states. 

\paragraph{Parameters} To obtain agent's policy $1000$ iterations as described in section \ref{iterations} are performed before starting simulations. In the performed simulations $\alpha$ was varied from $1$ to $10$ in steps of $1$, and $\beta=10/\alpha$. Each simulation was ran for $25000$ timesteps, divided into $10$ intervals of $2500$ timesteps, for each of which the measures under consideration (part of time spent in the corridor) are obtained, and then we report their average and  error bars based on the standard error from those measurements.

\subsection{Iteration scheme} 
\label{iterations}

Rather than using gradient descent directly of the $V$ over Eq. (\ref{eq:L_dual_final}), we developed an iteration scheme to speed up the search for the optimal policy. We restrict ourselves to the case of environments with deterministic state transitions. We empirically show that the algorithm finds a critical point of the dual function, and thus it provides us with the optimal $p^*(s,a)$.

Taking derivatives of the dual function in Eq. (\ref{eq:L_dual_final}) with respect to $V$ leads to the critical point condition (\ref{eq:critical_points_dual}). Defining $Q_s=\sum_{a}\exp(A_V(s,a)/\alpha)$, the derivative with respect to $V(s)$ can be written as 
\begin{equation}
\sum_{s'}(\sum_{a}p(s\left|s',a\right.)\exp(A_V(s',a)/\alpha))
Q_{s'}^{\alpha/\beta-1}-Q_{s}^{\alpha/\beta}=0
\label{eq:Q_opt}
\end{equation}
\noindent

Now, we consider a MDP where transitions are deterministic and rewards are zero everywhere ($r(s,a)=0$). Under the deterministic assumption, there is a one-to-one mapping between accessible state $s$ and action $a$ from state $s'$, that is, $s=s(s',a)$. Therefore Eq. (\ref{eq:Q_opt}) can be expressed as
\begin{equation}
Q_{s}^{\alpha/\beta} = 
\sum_{s'}
w_{s s'} e^{(V(s)-V(s'))/\alpha}
Q_{s'}^{\alpha/\beta-1}
\label{eq:derivative_0_2}
\end{equation}
\noindent
where we have used the notation $w_{s s'} = p(s|s',a) \in \{0, 1\}$ to indicate that the sum over $s'$ will only include terms from where $s$ is a possible successor state from $s'$, and $A_V(s',a)=V(s) - V(s')$ due to the deterministic transition from $s'$ to $s$ after performing the associated action $a$.

Introducing the notation $z_s=\exp(V(s)/\alpha)$, we first note that $Q_s=\sum_{a}\exp(A_V(s,a)/\alpha)$ can be written as $Q_s = \sum_{s'} w_{s' s } z_{s'}/z_s$, and that Eq. (\ref{eq:derivative_0_2}) becomes $Q_{s}^{\alpha/\beta} = z_s \sum_{s'} w_{s s'} z_{s'}^{-1} Q_{s'}^{\alpha/\beta-1}$. Inserting $Q_s$ from the first equation into the second, and solving for $z_s$ leads to 
\be
z_s = \left( \frac{(\sum_{s'} w_{s' s} z_{s'})^{\alpha/\beta}}{\sum_{s'} w_{s s'} z_{s'}^{-1} Q_{s'}^{\alpha/\beta-1}} \right)^{\beta/(\alpha+\beta)}
\label{eq:z_s}
\ee

Finally, we transform this fix point condition into an iterative scheme ($n=0,1,...$), 

\bea
&& Q_{s}^{(n)} = z_{s}^{{(n)}\; -1} \sum_{s'} w_{s' s } z_{s'}^{(n)}
\nonumber
\\
&& z_s^{(n+1)} = \left( \frac{(\sum_{s'} w_{s' s} z_{s'}^{(n)})^{\alpha/\beta}}{\sum_{s'} w_{s s'} z_{s'}^{{(n)} -1} Q_{s'}^{(n) \; \alpha/\beta-1}} \right)^{ \beta/(\alpha+\beta)}
\;.
\label{eq:scheme1}
\eea

As initial conditions we use $z_s^{0}=1$, except for absorbing states, for which we use $z_s^{0}=0$. It is possible to see that for absorbing states, defined as states where a transition occurs with certainty to itself, we have that $z_s^{n}=0$ for all $n$. Indeed, the $0$ value is stable over iterations: for $n=1$ we have $\sum_{s'} w_{s' s} z_{s'}^{0} = w_{s s} z_s^{0} = 1 \times 0 = 0$ in the numerator Eq. (\ref{eq:scheme1}), and so on for $n>1$.
In addition, for absorbing states, the value has to be minus infinity, and thus $z_s=0$.
To see this, note that $A_V(s,a)=V(s)-V(s)=0$ for these states, and therefore using Eq. (\ref{eq:Q_opt})  $\sum_{s'}\exp(A_V(s',a)/\alpha)Q_{s'}^{\alpha/\beta-1}-Q_{s}^{\alpha/\beta}=\sum_{s'\neq s}\exp(A_V(s',a)/\alpha)Q_{s'}^{\alpha/\beta-1}>0$,
so that Eq. (\ref{eq:Q_opt}) cannot be satisfied by any finite $V_{s}$, and thus we should have $V(s) \rightarrow-\infty$, and hence $z_{s}=0$ for absorbing states.

We empirically observe that for values $\alpha \leq \beta$ the scheme converges to the optimal solution, but we also observe that for $\alpha>\beta$ the scheme diverges. This instability is caused by the negative power $(1-\alpha/\beta)\frac{\beta}{\alpha+\beta}$ of $z^{(n)}_s$, because $z^{(n)}_s$ is contained in $Q^{(n)}_{s'}$ in the sum of the denominator on the right side of Eq. (\ref{eq:scheme1}).
To avoid this, we multiply both sides of Eq. (\ref{eq:z_s}) by $z_{s}^{\frac{\alpha-\beta}{\alpha+\beta}}$ and solve for $z_s$ to obtain the new iteration scheme  

\be
z_s^{(n+1)} = \left( 
\frac{
z_{s}^{(n) \; -1 + \alpha/\beta}
(\sum_{s'} w_{s' s} z_{s'}^{(n)})^{\alpha/\beta}}
{\sum_{s'} w_{s s'} z_{s'}^{{(n)} -1} Q_{s'}^{(n) \; \alpha/\beta-1}} \right)^{ \beta/2\alpha}
\;.
\label{eq:scheme2}
\ee

For $\alpha=\beta$ both schemes coincide, providing a smooth transition
between them.

\end{document}